%% file: iclr2024_conference.tex
\documentclass{article} %
\usepackage[usenames,dvipsnames,svgnames,table]{xcolor}
\usepackage{iclr2024_conference,times} 
\input{math_commands.tex}

\usepackage[colorlinks,allcolors=MidnightBlue]{hyperref}
\hypersetup{citecolor=OliveGreen,linkcolor=BrickRed}
\usepackage{url}
\usepackage{multirow}
\usepackage{booktabs}
\usepackage{arydshln}
\usepackage{graphicx}
\usepackage{subfigure} 
\usepackage{tipa}
\usepackage{wrapfig, blindtext}
\usepackage{pifont}
\usepackage[draft]{minted}
\usepackage{soul}
\usepackage{color}
\usepackage{amssymb}

\DeclareRobustCommand{\hllime}[1]{{\sethlcolor{lime}\hl{#1}}}

\newcommand{\our}[0]{\texttt{SCREWS}}

\newcommand{\x}{\texttt{x}}

\title{ \our\ \includegraphics[height=1.5em]{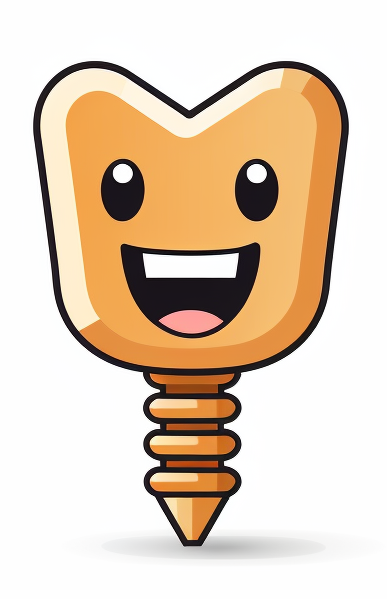} :  A Modular Framework for \\ Reasoning with Revisions}

\iclrfinalcopy
\author{%
  Kumar Shridhar \thanks{Work done while interning at Microsoft.}\ $\ ^{\dag}$ \quad Harsh Jhamtani $^{\ddag}$ \quad   Hao Fang $^{\ddag}$ \\  \bf Benjamin Van Durme $^{\ddag}$ \quad Jason Eisner $^{\ddag}$ \quad Patrick Xia $^{\ddag}$ \\
  $^{\dag}$ ETH Zurich \quad $^{\ddag}$ Microsoft Semantic Machines\\
  \texttt{shkumar@ethz.ch}\quad \texttt{patrickxia@microsoft.com}
}

\begin{document}

\maketitle

\begin{abstract}
Large language models (LLMs) can improve their accuracy on various tasks through iteratively refining and revising their output based on feedback. We observe that these \textit{revisions} can introduce errors, in which case it is better to roll back to a previous result. Further, revisions are typically homogeneous: they use the same reasoning method that produced the initial answer, which may not correct errors.
To enable exploration in this space, we present \our, a modular framework for reasoning with revisions.
It is comprised of three main modules: \emph{Sampling}, \emph{Conditional Resampling}, and \emph{Selection}, each consisting of sub-modules that can be hand-selected per task. We show that \our\ not only unifies several previous approaches under a common framework, but also reveals several novel strategies for identifying improved reasoning chains. We evaluate our framework with state-of-the-art LLMs (ChatGPT and GPT-4) on a diverse set of reasoning tasks and uncover useful new reasoning strategies for each: arithmetic word problems, multi-hop question answering, and code debugging. %
Heterogeneous revision strategies prove to be important, as does selection between original and revised candidates.
\end{abstract}

\section{Introduction}
\begin{figure}[t]
    \centering
    \includegraphics[width=0.86\textwidth]{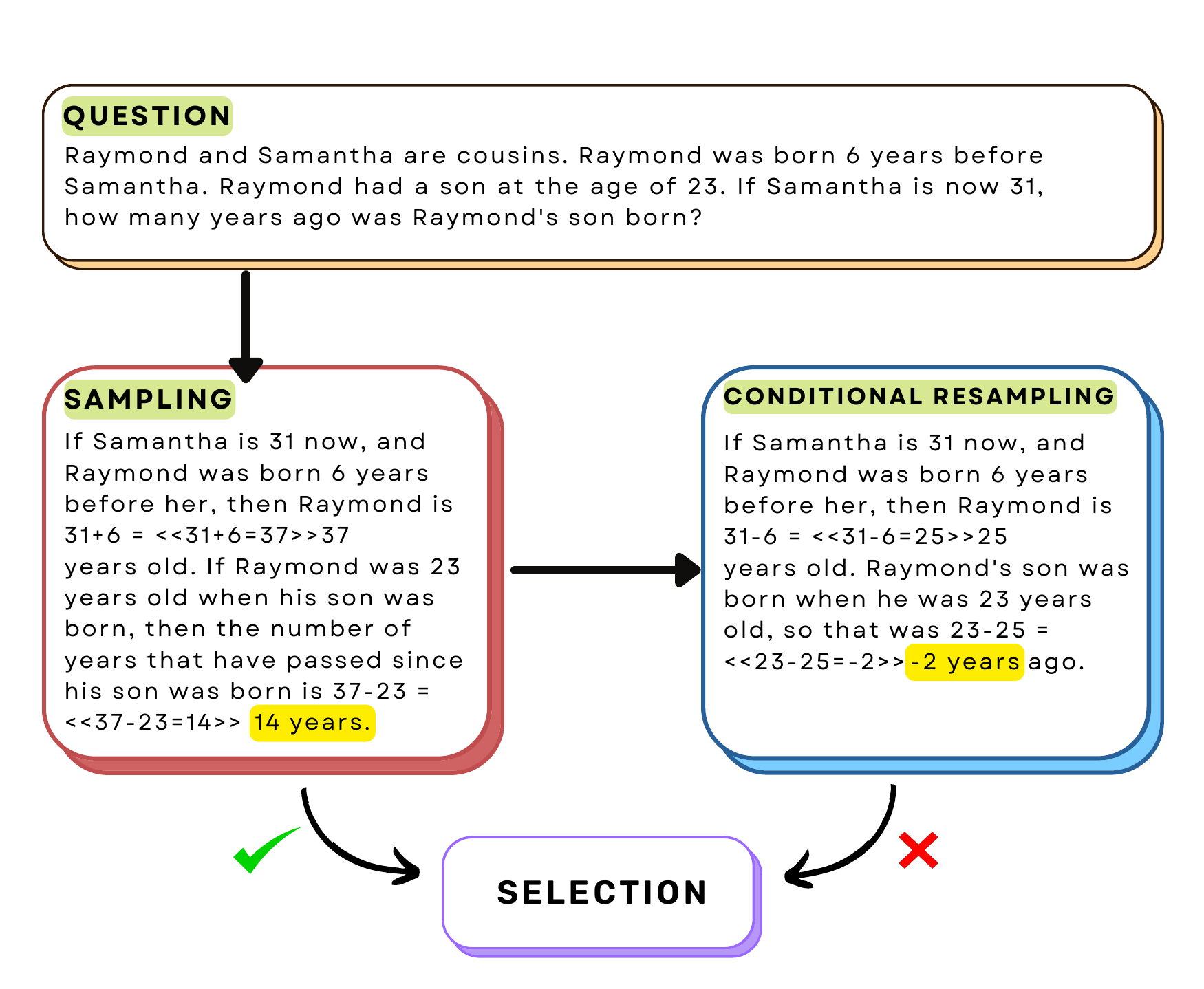}
    \caption{An example demonstrating that \emph{Conditional Resampling} (also known as ``\textit{refinement}'') can lead to incorrect modification of the original answer.  A \emph{Selection} module can decide to retract the modification and instead choose the original answer, which in this case is the correct one.}
    \label{fig:samp-resamp}
\end{figure}

Large Language Models (LLMs) have proven effective on a variety of reasoning tasks \citep{openai2023gpt4}. 
However, the LLM output is not always correct on its first attempt, and it is often necessary to iteratively refine the outputs to ensure that the desired goal is achieved \citep{self-refine, self_correct, php}. These refinement methods assume that subsequent outputs (either by the same model, or by an external model or some tool) lead to better performance. However, there is no guarantee that subsequent versions must be better; as Figure \ref{fig:samp-resamp} illustrates, refinement can lead to a wrong answer. This motivates a \textit{Selection} strategy whereby the model can select an earlier output. 

In addition, past work on iterative refinement typically assumes a single, fixed reasoning strategy \citep{self_correct, self_improve, self-refine, php}. Humans, however, are more flexible.  A student preparing for an exam may use deductive reasoning to solve problems and inductive reasoning to verify the results; or a product manager may use a brainstorming strategy to list several ideas and then switch to a prioritization strategy to rank them based on their feasibility or impact. 
Thus, we propose a \textit{modular} approach to answer refinements, allowing us to test different strategies. 

\begin{figure}[t]
    \centering
    \includegraphics[width=1\textwidth]{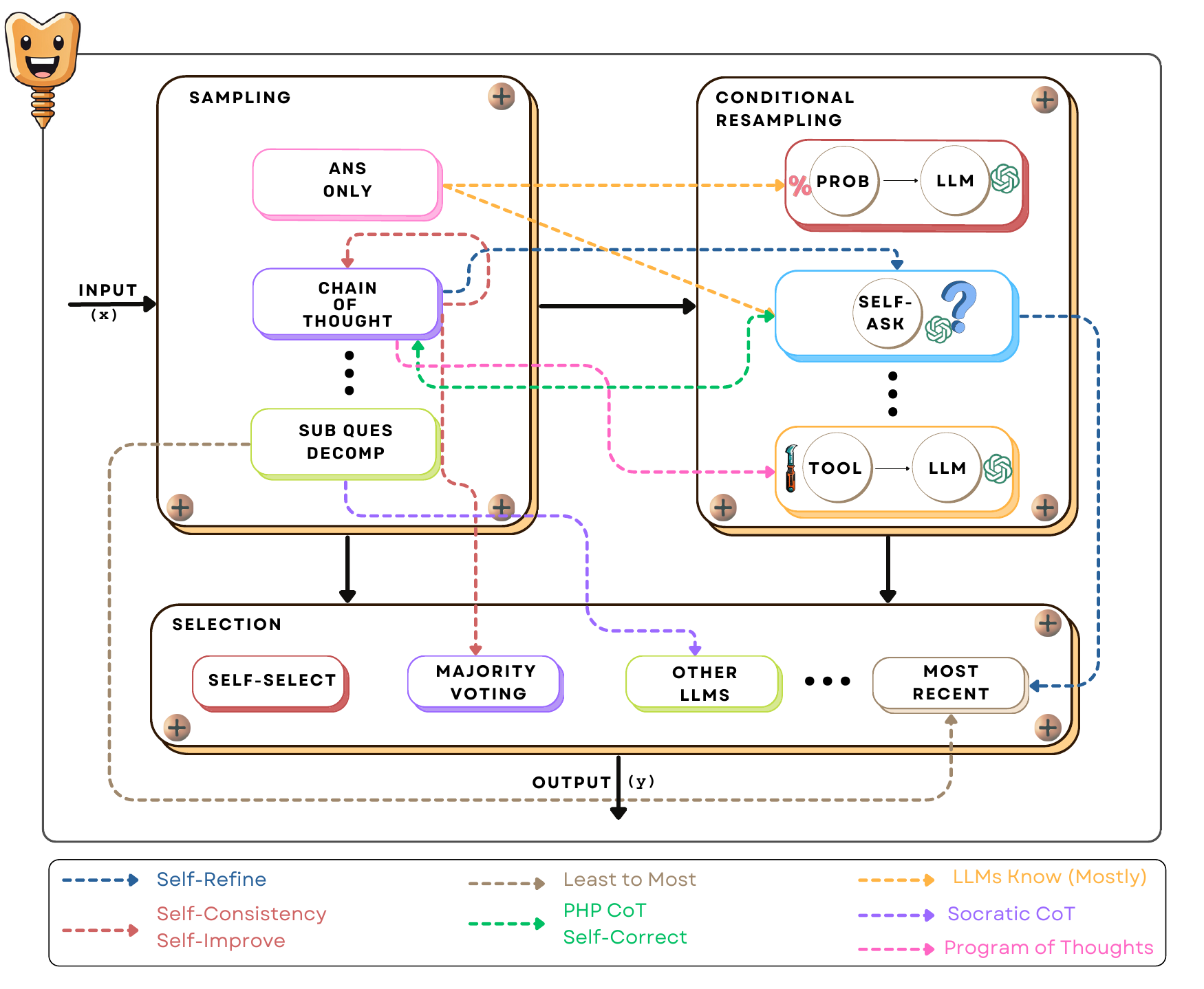}
    \caption{Overview of our modular framework for reasoning with revisions, \our. Each of the three large boxes (``modules'') contains several alternatives (``submodules'').  A lot of past works can be viewed as instances of our framework, namely Self-Refine \citep{self-refine}, Least to Most \citep{leasttomost}, LLMs Know (Mostly) \citep{kadavath2022language}, Self-Consistency \citep{selfconsistency}, Self-Improve \citep{self_improve}, PHP CoT \citep{php}, Self-Correct \citep{self_correct}, Socratic CoT \citep{socratic_cot}, Program of Thoughts \citep{programofthoughts}, among many others. (...) represents other sub-components that can be added to each module, like cached memory or web search for \textit{Sampling}, fine-tuned model or external verifier for \textit{Conditional Resampling}, and human- or oracle-based selection for the \textit{Selection} module, among others.}
    \label{fig:screws}
\end{figure}

In this work, we introduce \our, a modular framework for reasoning with revisions.\footnote{\our\ \includegraphics[height=1.3em]{images/screws_logo.png} stands for ``\textbf{S}ampling, \textbf{C}onditional \textbf{RE}sampling \textbf{W}ith \textbf{S}election.'' Our code and results are available at  \url{https://github.com/kumar-shridhar/Screws/}.}  Figure \ref{fig:screws} introduces the three main modules of the framework in detail, namely \textit{Sampling}, \textit{Conditional Resampling}, and \textit{Selection}. For a given task and input sequence, we instantiate \our\ by fixing the submodules for each module (for example, we might select ``Chain of Thought'' for \textit{Sampling}). The initial outputs generated by \textit{Sampling} are passed to \textit{Conditional Resampling}, which decides whether to generate a revision \textit{conditioned} on the initial sample, and does so if needed. Finally, all samples and revisions are given to the \textit{Selection} module, which selects the best one. Given the modular nature of our framework, several recently proposed self-refining methods can be improved by using other components of the framework. An example is the combination of the self-refinement method \citep{self-refine} with our model-based selection strategy, which can improve overall performance; more such strategies are described in section \ref{results}. 

We evaluate \our\ on a variety of reasoning tasks: arithmetic reasoning, multi-hop question answering, and code debugging, using ChatGPT \citep{brown2020language} or GPT-4 \citep{openai2023gpt4}.  Our proposed strategies achieve substantial improvements (10--15\%) over vanilla strategies of sampling and resampling. We demonstrate the usefulness of heterogeneous resampling, which can help the model modify its reasoning, leading to a substantial improvement over the baselines at a very low overall cost. We also discuss the importance of a model-based selection strategy that allows the model to roll back to its previous more confident outputs, an important component for modern LLMs. 

\section{Background}
\paragraph{Sampling} Prompting LLMs to generate a series of intermediate steps has proven to be effective for improving their reasoning capabilities \citep{wei2023chainofthought, minerva, kojima2022large, selfconsistency}. Some approaches in this direction include Chain of Thought \citep{wei2023chainofthought, zhang2022automatic, selfconsistency} and adding ``Let's think step by step" to the prompt \citep{kojima2022large}. Another approach is ``question decomposition'', which decomposes the main problem into simpler problems and solves them iteratively \citep{min2019multi, socratic_cot, leasttomost, jhamtani2023natural, radhakrishnan2023question}. Each of these approaches has its own advantages depending on the underlying task \citep{shridhar-etal-2023-distilling}. However, we are not aware of work combining these methods. 

\paragraph{Conditional Resampling} 
The use of feedback to improve generated samples has been well studied, where the feedback can come either from humans \citep{tandon2021interscript, bai2022training, elgohary-etal-2021-nl}, from reward models \citep{ziegler2020finetuning, lu2022quark, socratic_cot, christiano2023deep, lightman2023lets}, from external tools such as code interpreters \citep{schick2023toolformer, programofthoughts}, or from other LLMs \citep{self-refine, self_correct, fu2023gptscore, peng2023check, yang-etal-2022-re3, php, Cohen_Hamri_Geva_Globerson_2023, Ling_Fang_Li_Huang_Lee_Memisevic_Su_2023, khalifa2023discriminatorguided}. However, even if these feedback mechanisms are infallible, the resulting revisions may introduce new errors.\footnote{Prior work uses the term ``refinement,'' which we do not use because refinement implies finer (improved) responses, which does not always occur.}

\paragraph{Selection} 
When using LLMs to evaluate and revise the output, the most common selection technique is to always select the final output  \citep{self-refine, shinn2023reflexion, php, yao2023react, chen2023teaching, weng2023large}. However, this can lead to accepting incorrect changes made to previously correct outputs. Other selection methods involve ranking multiple sampled outputs \citep{GSM8k} or majority voting \citep{selfconsistency, minerva, php}. These methods often use a homogeneous sampling strategy with changes in temperature or other similar hyper-parameters. Our work extends the strategy to heterogeneous sampling and selection.

\section{\our: Methodology}

In this section, we describe \our, our proposed modular framework for reasoning with revisions to tackle different reasoning tasks. Given a problem \texttt{x}, the goal is to generate an \textit{answer} $a$, which in our experiments may be a string or a number.
\our\ consists of three main modules: \textit{Sampling}, \textit{Conditional Resampling}, and \textit{Selection}.  Different variants of \our\ are obtained by instantiating these modules in different ways.  The options for each module are described below and illustrated schematically in Figure \ref{fig:screws}. 

All of our methods will invoke one or more stochastic functions, where each function $\psi$ maps a tuple of input strings to a \textit{result} string \texttt{y} that contains useful information.  In practice, $\psi$ deterministically constructs a prompt from the input strings and then samples \texttt{y} from a large pretrained language model as a stochastic continuation of this prompt.  For a given tuple of input strings, the prompt constructed for $\psi$ will typically be a formatted encoding of this tuple, preceded by a task specific instruction and several demonstrations (few-shot examples) that illustrate how $\psi$ should map other encoded input tuples to their corresponding continuations \citep{brown2020language}.
For concreteness, the prompts we use in our experiments are illustrated in Appendix \ref{app:prompts}.\looseness=-1

\subsection{Sampling}\label{sec:sampling}
We consider three instantiations of the sampling module.  Different instantiations may be appropriate for different tasks.

\paragraph{Answer Only} In this method, for a given problem \texttt{x}, the model $\psi$ directly generates the answer $\texttt{y} = \psi(\texttt{x})$ without any intermediate steps. This is the simplest and most naive sampling method. The value of \texttt{y} is returned as the answer $a$ (if there is no further revision of $\texttt{y}$).

\paragraph{Chain of Thought (CoT)} For many reasoning tasks today, generating explanations improves the quality of the final answer \citep{wei2023chainofthought, kojima2022large}. Chain of Thought sampling encourages the model to explain the intermediate step-by-step reasoning en route to a decision. This approach is now commonly used in several reasoning tasks. Again, we define $\texttt{y} = \psi(\texttt{x})$, but now we expect the prompt continuation to consist of step-by-step reasoning culminating in the step by step answer $\texttt{y}$, as demonstrated by the few-shot examples included in the prompt. The answer $a$ is extracted from \texttt{y} using a simple deterministic pattern-matching heuristic.

\paragraph{Sub-question decomposition} This method decomposes the problem \texttt{x} into simpler sub-questions $[x_1, x_2, \dots , x_n]$. For each sub-question $x_i$ in turn ($i=1,2,\dots,n$), the model is called to generate the corresponding sub-answer $y_i = \psi(\texttt{x},x_1,y_1,\ldots,x_{i-1},y_{i-1},x_i)$.  
Note that we generate all questions before seeing any answers; that choice follows \citet{shridhar-etal-2023-distilling}, who found this approach to work better than interleaved generation of questions and answers.  The sequence of questions may be generated in a single step, either by a call to a stochastic function $\psi_{\text{question}}$, or by a custom question generation module that has been fine-tuned on human-written questions as in \citet{GSM8k}. The answer $a$ is extracted from $y_n$ with a simple heuristic as in CoT.

\subsection{Conditional Resampling}
\label{conditional_resampling}
The result $\texttt{y}$ from the \textit{Sampling} module can be viewed as a \textit{provisional result}, \texttt{y}\textsubscript{curr}. This is passed to the \textit{Conditional Resampling} module where a decision is made whether or not to revise it. This is done in two steps: first deciding whether or not to revise, and then if so, resampling a new result $\texttt{y}\textsubscript{next}$ using one of the sampling methods mentioned above. The resampling is conditional because $\texttt{y}\textsubscript{next}$ may depend on  $\texttt{y}\textsubscript{curr}$.  While there are many methods for \textit{Conditional Resampling}, our work focuses on the following instantiations:

\paragraph{Self-Ask}  \citet{kadavath2022language} 
uses a function $\psi\textsubscript{ask}(\texttt{x}, \texttt{y}\textsubscript{curr})$.  The first token of the result indicates whether $\texttt{y}\textsubscript{curr}$ is correct, for example by starting with ``Yes'' or ``No''.  If ``Yes'', we do not resample; if ``No'', we must resample a revised answer $\texttt{y}\textsubscript{next}$.  In principle, the revision could be iterated, although \citet{kadavath2022language} did not do this, nor do our experiments in this paper.

In our version of self-ask,  $\psi\textsubscript{ask}$ is formulated so that $\texttt{y}\textsubscript{next}$ appears in the result string $\psi\textsubscript{ask}(\texttt{x}, \texttt{y}\textsubscript{curr})$ following the word ``No''.  Thus, both steps are efficiently performed by a single call to $\psi\textsubscript{ask}(\texttt{x}, \texttt{y}\textsubscript{curr})$.  For this method, we always use greedy decoding (temperature 0), which deterministically selects whichever of ``Yes'' or ``No'' is more probable.\footnote{A threshold other than 50\% could be tuned to optimize the downstream reward of the whole system.  This compensates for bias toward the ``Yes'' or ``No'' token, and also considers how much resampling followed by selection will actually improve the final accuracy and harm the speed of the system.
Orthogonally, the correctness probability of \texttt{y}\textsubscript{curr} could be assessed by a dedicated $\psi_{\text{check}}(\texttt{x}, \texttt{y}\textsubscript{curr})$, but we were unsuccessful with this as $\psi_\text{check}$ was poorly calibrated, mirroring findings on model calibration \citep{kadavath2022language, Xiong_Hu_Lu_Li_Fu_He_Hooi_2023}.\looseness=-1}  Demonstrations for the prompt are shown in Appendix \ref{resampling_prompt}.

When the sampling module (Section~\ref{sec:sampling}) used sub-question decomposition to produce a chain of sub-answers $\texttt{y}\textsubscript{curr}=[y_1,\ldots,y_n]$, rather than checking and revising only the final result step $y_n$ by calling $\psi_{\text{ask}}(\texttt{x}, y_{n})$, we can instead check and revise each step, at the cost of more calls to $\psi_{\text{ask}}$.  For each provisional sub-answer $y_i$ in turn (starting with $i=1$), we predict whether it is correct by calling $\psi_{\text{ask}}(\texttt{x},x_1,y_1,\ldots,x_{i-1},y_{i-1},x_i, y_i)$. The first time the output is ``No'', we resample $y'_i$ through $y'_n$, yielding the revised result $\texttt{y}_{\text{next}} = [y_1, \ldots, y_{i-1}, y'_i, \ldots, y'_n]$. In principle, self-ask could then be applied again at later steps $> i$ of both the original and revised chains; then choosing among the many resulting chains, using the selection procedures of the next section, would resemble branching in a reasoning tree \citep{yao2023tree}.

\paragraph{Tool-Based LLM} For some tasks, we construct $\psi\textsubscript{ask}$ so that it is allowed to use tools \citep{schick2023toolformer}.  The reason is that in tasks like fact-checking, it is futile to ask the LLM to check $\texttt{y}_\text{curr}$ because it might not have the requisite knowledge for evaluation. 
The tools can be used to collect additional information or facts to help the model detect and fix problems in its own generated answer.  Tools like search engines or fact retrievers can be used to evaluate correctness and generate a new revision.  Some other tools like code interpreters are not capable of generating text, but can still be used to evaluate correctness.

\subsection{Selection}
The last module in \our\ is the \textit{Selection} module. 
In this step, we use either a model $\psi\textsubscript{select}$ or simple heuristics to select the \emph{final}
result \texttt{y} from which we then extract the \emph{final} answer $a$.
In effect, this allows us to construct a simple ensemble of multiple systems.

\paragraph{LLM-Based Selection} Just as an LLM was used above to evaluate whether $\texttt{y}\textsubscript{curr}$ is good, an LLM can be used to evaluate whether $\texttt{y}\textsubscript{next}$ is better.
We call $\psi\textsubscript{select}(\x, \texttt{y}\textsubscript{curr},\texttt{y}\textsubscript{next})$ to choose between two result strings.\footnote{We found that the order of $\texttt{y}\textsubscript{curr}$ and $\texttt{y}\textsubscript{next}$ in the prompt was unimportant; in our reported results, we randomized this order.}  
Note that it could be naturally extended to choose among more than two answers. When selection and sampling are implemented using the same LLM, we refer to the method as \emph{self-select} (e.g., in Figure \ref{fig:screws}). The prompts for $\psi\textsubscript{select}$ in our experiments are shown in Appendix \ref{selection_prompt}. %

\paragraph{Rule-Based Selection} We consider the other methods we study to be rule-based. Past work on iterative refinement \citep{self-refine, self_improve, php} always selects the most recent revision. Majority voting is a simple traditional ensembling method that has been used for selection
\citep{selfconsistency, minerva}, but it is costly because it requires several samples. 

\subsection{Other Possibilities} There are other possible ways to instantiate each module.  Tools like web-based search or cache-based retrieval could be used to generate the initial attempt in the \textit{Sampling} module.  A fine-tuned classification model could be used to verify outputs in the \textit{Conditional Resampling} module.  Similarly, a fine-tuned model could be used for the \textit{Selection} module. In this paper, however, we study only the instantiations described above.

\section{Experiments}
\subsection{Tasks}

We test the effectiveness and flexibility of \our\ on three categories of reasoning tasks: GSM8K \citep{GSM8k} for arithmetic reasoning, StrategyQA \citep{strategyqa} for multi-hop question answering, and Big-Bench \citep{big-bench} AutoDebugging\footnote{\url{https://github.com/google/BIG-bench/tree/main/bigbench/benchmark_tasks/auto_debugging/}}  for code debugging. The GSM8K dataset is a grade-school-level math word problem  dataset with a test set of 1319 samples, each requiring two to eight steps to solve. GSM8K includes sub-questions that were generated by a fine-tuned GPT-3 model
and correspond to the steps in a particular correct CoT solution. 
Since these sub-questions were generated with oracle knowledge of a correct CoT solution, we refer to experiments using them as``Subq (Or)''.  We use  ``Subq (QG)'' for the fairer experimental condition where we instead generated the subquestions from ChatGPT using two-shot prompts (which are provided in Appendix \ref{qg_prompt}).\footnote{Unsurprisingly, the Subq (Or) sub-questions proved to be consistently better, as we will see in Section~\ref{results}. In addition to their oracle knowledge of a human-written answer, some of the sub-questions themselves may also have been human-written: the sub-question generation model was fine-tuned on around 800 human-written examples, and some of those examples may also be included in the released dataset (\url{https://github.com/openai/grade-school-math\#socratic-dataset}).}

Following \citet{magister2023teaching} and \citet{shridhar-etal-2023-distilling}, we test on the first 490 samples from the training set of StrategyQA (since their test set is unlabeled). The demonstration examples for our various stochastic functions $\psi$ were drawn randomly from the rest of the training set.
StrategyQA also includes human-annotated oracle subquestions (which we again use for ``Subq (Or)'' results) and related facts that can assist in answering the main question (which we use for tool-based conditional resampling as in Section~\ref{conditional_resampling}). Finally, the Auto Debugging dataset tests whether
a model can answer questions about the intermediate state of a program without executing the code. The dataset consists of 34 coding examples, of which 33 were used as test examples and 1 as a demonstration example in the prompt.

\subsection{Experimental Setup}

We always report exact-match accuracy: the percentage of examples on which our final answer $a$ matches the gold answer.
For all of our experiments, we use the ChatGPT API \citep{brown2020language} from July 2023 (\texttt{gpt-3.5-turbo-0301}).  This model is a decoder-only Transformer LLM \citep{vaswani2017attention} that was fine-tuned using reinforcement learning with human feedback \citep{ziegler2020finetuning, christiano2023deep}. Some experiments were also performed using GPT-4 \citep{openai2023gpt4} to show the scaling capabilities of our framework. 

\paragraph{Sampling} With all choices of the \textit{Sampling} module, we use 5-shot sampling for GSM8K and StrategyQA and 1-shot sampling for Auto Debugging. Greedy decoding (temp = 0) is used for the main experiments while higher temperature (0.7) is used for the majority voting experiments (one sample was generated with temp = 0 and the other four at temp = 0.7). All prompts are provided in Appendix \ref{sampling_prompts}.

\paragraph{Conditional Resampling} Greedy decoding is used to first make a binary resampling decision and then to sample. 4-shot prompts (with two correct and two incorrect samples) are used for the GSM8K and StrategyQA datasets, while a 2-shot prompt (with one correct and one incorrect sample) is used for Auto Debugging. 
For StrategyQA, we use tool-based resampling by including the provided facts from the dataset into the prompt (Appendix \ref{resampling_prompt}) to simulate a (perfect) fact retrieval tool. %

\paragraph{Selection} For the \textit{self-select} strategy, the prompts include two examples and selection was produced with greedy decoding (prompts in  Appendix \ref{selection_prompt}). For majority voting, a majority vote on the final answers was taken over $k \in \{1,3,4,5\}$ samples. Ties were broken randomly.

\section{Results}
\label{results}

\begin{table*}[t]
\centering
\begin{tabular}{ccc}
\toprule 
{\bf Sampling} & {\bf Conditional Resampling} & {\bf Accuracy} \\
\midrule
\multirow{4}{*}{CoT} & - & 71.64 \\
& CoT & 73.00 \\
& Subq (QG) & 73.69 \\
& Subq (Or) & \textbf{73.99} \\
\midrule
\multirow{3}{*}{Subq (QG)} & - & 71.87 \\
& CoT & \textbf{73.99} \\
& Subq (QG) & 71.26 \\
\midrule
\multirow{3}{*}{Subq (Or)} & - & 78.62 \\
 & CoT & \textbf{78.99} \\
& Subq (Or) & 78.24 \\
\bottomrule 
\end{tabular}
\caption{The improvements achieved by using \emph{Conditional Resampling} for the GSM8K dataset, where $\texttt{y}_\text{next}$ is always selected. \textbf{CoT} refers to the Chain of Thought method, while \textbf{Subq} refers to the Subquestion Decomposition method. \textbf{Subq (QG)} refers to the case where subquestions are generated by the ChatGPT model, while \textbf{Subq (Or)} refers to the Oracle questions present in the Socratic version of the dataset.}
\label{ResultsGSM}
\end{table*}

\subsection{GSM8K}
\label{results:gsm8k}
\paragraph{Conditional Resampling Works Better with Method Change} Previous work \citep{self-refine} has shown that when a chain-of-thought method is used for initial \emph{Sampling}, reasoning ability is improved by \emph{Conditional Resampling} with the same method.  The benefit comes from taking the previous sample into account. 

We reproduced this previous finding: the CoT scores for GSM8K improved by 1.4 points after resampling with CoT (71.6 to 73.0), as shown in Table \ref{ResultsGSM}. However, when the initial \textit{Sampling} used subquestion decomposition,  we found that resampling with subquestion decomposition actually harmed accuracy.  It decreased the score by about 0.5 points (71.9 to 71.3 with generated subquestions, 78.6 to 78.2 with oracle subquestions). 

What gave the best results---for all three \textit{Sampling} methods---was \textit{Conditional Resampling} with a \emph{different} method from the originally chosen one.  It gave a large gain over Sampling when the original Sampling used CoT and Resampling used subquestion decomposition (71.6 to 73.7, with generated subquestions) and vice versa (71.9 to 74.0).
Even with oracle subquestions, moderate gains are still seen when resampling with CoT (78.6 to 79.0). This demonstrates that it is useful to change methods using \textit{Conditional Resampling}, a novel finding with our framework.\footnote{In principle, we could also use resampling and selection to combine Subq (QG) with Subq (Or); we may try this in a future version of this paper.}

\begin{table*}[h!]
\small
\centering
\addtolength{\tabcolsep}{-1pt}
\begin{tabular}{l   ccc  ccc }
 \toprule 
\bf Method  &   \multicolumn{3}{c}{\bf Independent Sampling} & \multicolumn{3}{c}{\bf Conditional Resampling}  \\
\cmidrule(lr){2-4}
\cmidrule(lr){5-7}
{} & CoT &  Subq (QG) & Subq (Or) & CoT  & Subq (QG) & Subq (Or) \\
\hline
\addlinespace
CoT & 71.64   & 74.90 [85.36] &  \textbf{81.34} [89.08] & 72.93 [73.08]  & 73.76 [73.76] & \textbf{73.99} [73.99] \\  
Subq (QG) & \textbf{74.90} [85.36]  & 71.87 & - &  \textbf{73.99} [75.43]  & 72.40 [72.40] & -\\
Subq (Or) & \textbf{81.34} [89.08]   & - & 78.62 & 78.99 [81.50] &   -  & \textbf{79.22} [79.22] \\
\addlinespace
\bottomrule 
\end{tabular}
\addtolength{\tabcolsep}{1pt}
\caption{Impact of \textit{Selection} on the GSM8K data set on \textit{Independent Sampling} and \textit{Conditional Resampling}. The upper bound from using a \textit{Selection} oracle is given in square brackets.}
\label{ResultsSelection}
\end{table*}

\paragraph{Importance of Selection Module}  \emph{Conditional Resampling} does not invariably improve every output.  In fact, we saw in Table \ref{ResultsGSM} that for some settings, it may harm the output quality even on average.  This is why the \textit{Selection} module is useful---to detect and reject cases of harmful revisions.

First, as a starting point, the left half of Table \ref{ResultsSelection} considers using Selection only as an ensembling technique to combine the outputs of two \textit{independent} Sampling strategies.  (Note that this matrix is symmetric.)  Although CoT and subquestion decomposition are about equally good Sampling strategies (71.6 and 71.9), using a Selection module to select the better of the two achieves a 3-point gain (to 74.9).  Much larger gains (up to 85.4) are potentially available from improving Selection---the upper bound on performance (if Selection always chose the better option) is shown in square brackets.  This shows that the two Sampling strategies have largely complementary errors. A similar pattern applies when the subquestion decomposition method is permitted to use oracle subquestions, which improves performance across the board to 81.34.

The right half of Table \ref{ResultsSelection} shows \textit{Selection} between the \textit{Sampled} and \textit{Conditionally Resampled} predictions from Table \ref{ResultsGSM}. (This matrix is asymmetric.) For CoT, the results remain the same at 73.99, which is due to the fact that the upper bound is at 73.99, showing no room for further improvement. For other cases with subquestioning, we see an improvement of up to 1 point.  Finally, we observe that the \emph{Selection} module is far from perfect and has room for further improvement, as seen from the upper bounds.  A \emph{Selection} method ought to look at features of the two answers that turn out to be correlated with correctness, and we hypothesize that models fine-tuned specifically for \emph{Selection} may prove more effective than few-shot learning at identifying these features.

The right half of Table \ref{ResultsSelection} is the cheaper method, because we observe $\psi_\text{ask}$ resamples on only 5-15\% of the examples rather than all of them.  
A tradeoff between accuracy and cost is shown in Figure \ref{fig-sampcost}.

\begin{figure}[h!]
\centering
\subfigure[Impact of the number of samples on the accuracy when majority voting is used as a \emph{selection} method.]{ %
\includegraphics[width=0.45\textwidth]{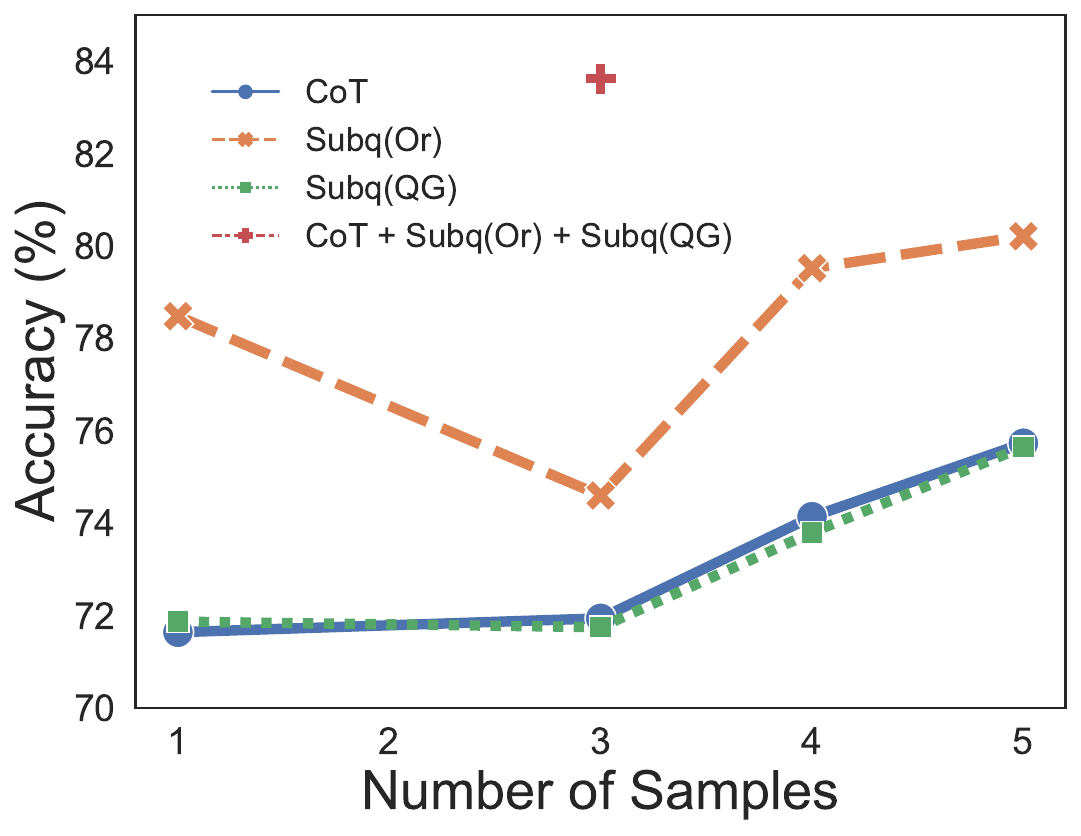} 
\label{fig:sub1} 
} 
\hfill 
\subfigure[Comparison of perfect selection (``+ perfect'') vs.\@ majority voting (``+ maj'') for different methods of CoT and Subq.]{ %
\includegraphics[width=0.50\textwidth]{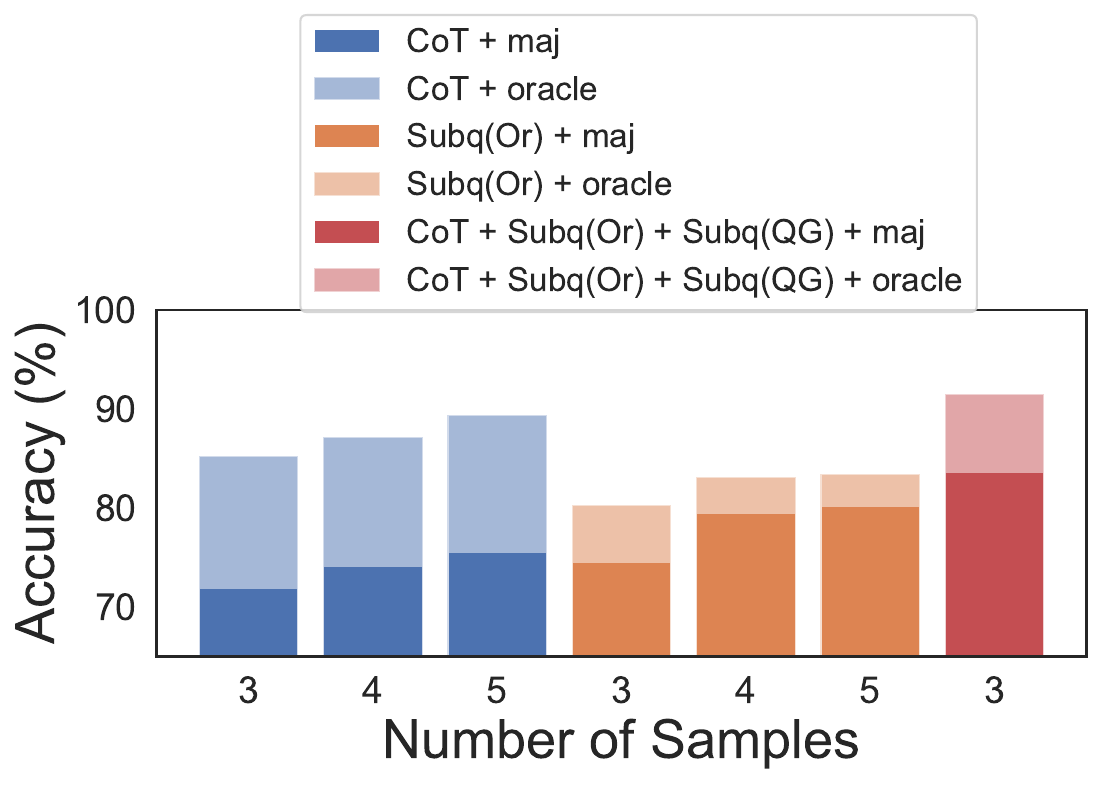} 
\label{fig:sub2} 
} 
\caption{The + in graph (a) shows that majority voting with 3 diverse samples (CoT + Subq(Or) + Subq(QG)) outperforms both CoT and Subq(Or) even with 5 samples. Graph (b) shows the potential of the \emph{selection} method when a perfect selector is used. It can be thought of as the upper bound of the selection mechanism. Both figures are for the GSM8K dataset.} 
\label{fig:accvssampl} 
\end{figure} 

\paragraph{Selection and Voting} Unweighted majority vote has been one of the most popular \textit{Selection} methods in past work \citep{selfconsistency, minerva, php}, since it requires no training. The two lines in Figure \ref{fig:accvssampl}(a) generally show improvement from \textit{Sampling} more times from the same model (at temperature 0.7) and \textit{Selecting} by majority vote. 

Recalling that the left half of Table \ref{ResultsSelection} showed benefit from ensembling independent samples from 2 different \textit{Sampling} methods (up to 81.34 accuracy when oracle subquestions are allowed), we observe that majority vote is a convenient way to do so for 3 different methods (where all methods can now use temperature 0).  This achieves 83.62 accuracy, as shown by the $\star$ in Figure \ref{fig:accvssampl}(a).  Of course, model-based \textit{Selection} could potentially do even better than majority voting.  The 7 points for $k \geq 3$ in (a) are repeated as the dark bars in Figure \ref{fig:accvssampl}(b), with the light bars showing the upper bounds that could be achieved by replacing majority voting with a perfect \textit{Selection} method.  The best upper bound corresponds again to the use of 3 different methods. In principle, one could ensemble over a larger set by allowing each of the 3 methods to contribute multiple samples.

\subsection{StrategyQA}

\paragraph{Vanilla resampling does not improve what model does not know $\rightarrow$ A need for tools} For the StrategyQA dataset, we observe in Table \ref{ResultsSQA} that accuracy is harmed by \emph{Conditional Resampling} with the same \textit{Sampling} method, without \textit{Selection}, as was sometimes the case for GSM8K.  On StrategyQA, however, even \textit{Selection} usually does not repair the problem, perhaps because StrategyQA requires multi-hop question answering.  When the model lacks the necessary factual knowledge, Self-Ask will be insufficient. A real example at the bottom of Figure \ref{fig:qualitative} shows how resampling can preserve an incorrect claim generated by the model. 

To help the model decide whether and how to revise the answer, we try including relevant facts (provided by StrategyQA) into the resampling prompt, as shown in Appendix \ref{prompt:sqa_with_facts}, to simulate the result one may get by using an external tool like a fact retriever.  As Table \ref{ResultsSQA} shows, this yields a 2-point improvement (``Facts\textsubscript{re}'' vs.\@ ``Internal\textsubscript{s}'') over \textit{Sampling}, for both CoT and Subq (QG).

We assume that tool invocations are expensive, which is why we include facts only during \textit{Conditional Resampling}. In practice, the initial result is revised only 10--35\%
of the time, and therefore ``Facts'' does not need to invoke a tool call for every input example.\footnote{However, if the facts were included during \textit{Sampling}, the performance can increase beyond 90\%.}  To achieve this speedup, we do not include facts in the prompt when initially calling to $\psi_\text{ask}$ to decide whether to resample, but only when we actually generate $\texttt{y}_\text{next}$.

\begin{table*}[h!]
\centering
\begin{tabular}{l  c  cc  cc  }
 \toprule
\bf Method  & \bf Sampling &  \multicolumn{2}{c}{\bf Conditional Resampling} &
\multicolumn{2}{c}{\bf Selection} \\
\cmidrule(lr){3-4}
\cmidrule(lr){5-6}
{Knowledge Source:} & Internal\textsubscript{s} & Internal\textsubscript{re} & Facts\textsubscript{re} & Int\textsubscript{s} vs.\@ Int\textsubscript{re} & Int\textsubscript{s} vs.\@ Facts\textsubscript{re} \\
\midrule
\multicolumn{6}{c}{\bf StrategyQA} \\
\midrule
\addlinespace
CoT & 77.18 & 74.54 & \textbf{79.02} & 75.76 & 78.41  \\  
Subq (Or) & 85.91 &  78.97 & 84.69 &  85.30 & \textbf{86.30}   \\
Subq (QG) & 78.16 &  74.69 &  \textbf{80.40} &  78.78 & 80.00  \\ 
\addlinespace
\midrule
\multicolumn{6}{c}{\bf Code Debugging} \\
\midrule
Answer Only & 73.52 & 82.35 & & 88.23 [91.20] &\\
CoT & 70.58 & 73.52& & 73.52 [73.52]& \\
\hdashline
Answer Only + CoT & - & - & & 85.29 [88.23]&\\
\bottomrule 
\end{tabular}
\caption{Comparing different strategies for the StrategyQA (top) and Big Bench Code Debugging (bottom) datasets. For StrategyQA, external facts are provided to the model (``Facts'') versus relying on the model's internal capabilities (``Internal''). The numbers in square brackets indicate upper bound performance, assuming perfect selection. Subscripts ``s'' and ``re'' refer to Sampling and Resampling respectively.}
\label{ResultsSQA}
\end{table*}

\subsection{Code Debugging}

\paragraph{The effectiveness of \our}
For the code debugging task, we observed that the Answer Only method achieves similar scores to CoT,\footnote{We do not experiment with subquestion decomposition as subquestions are not part of this dataset.} as reported in the bottom half of Table \ref{ResultsSQA}, suggesting that no particular \textit{Sampling} method is superior on all datasets. However, we see the benefits of using \our, as we find that with Answer Only, adding \textit{Conditional Resampling} followed by \textit{Selection} leads to a performance boost of 15 points (from 73.52 to 88.23). While the dataset size limits our ability to make concrete conclusions, the findings here support the conclusions drawn on other datasets: \emph{Resampling} and \emph{Selection} lead to benefits and heterogenous sampling can prove effective.

\section{Additional Analysis}
\subsection{Total Cost} 
\begin{wrapfigure}{r}{7.4cm}\vspace{-5mm}
    \includegraphics[width=0.95\linewidth]{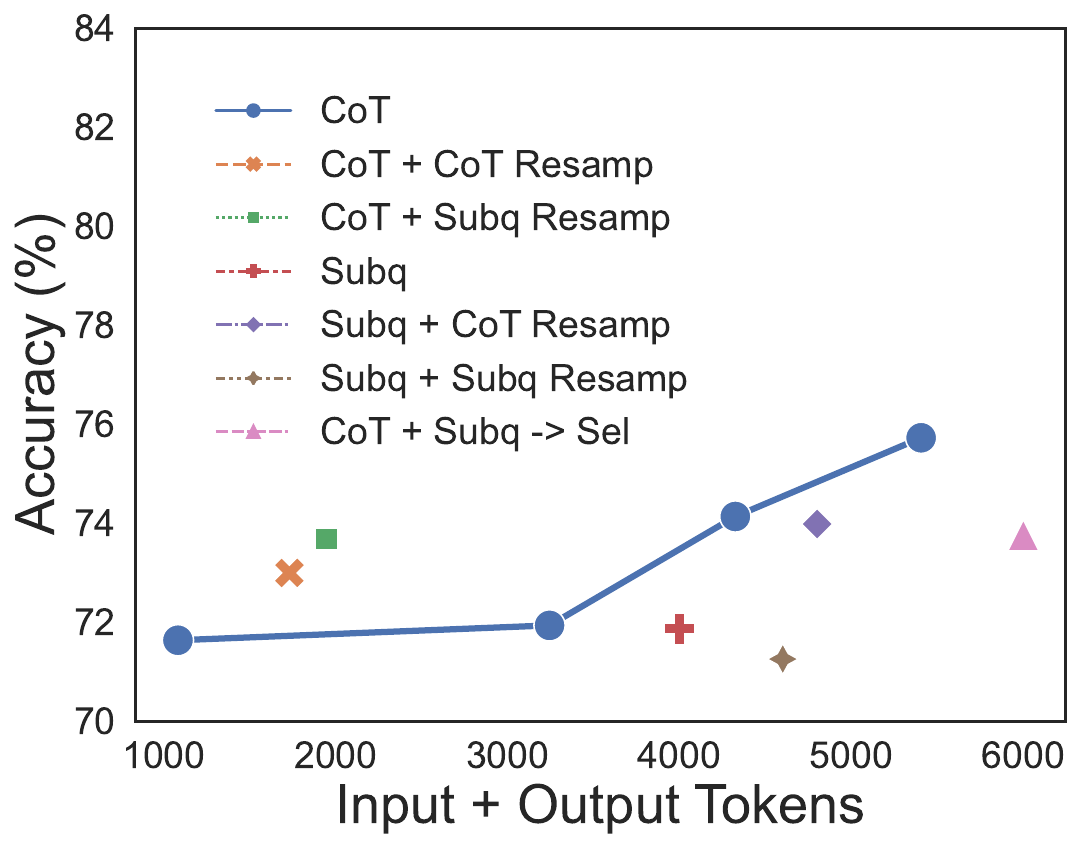}\vspace{-2mm}
    \caption{On GSM8K, sampling cost vs.\@ accuracy.  The blue line (copied from Figure~\ref{fig:sub1}) shows a baseline of majority voting over $k\in\{1,3,4,5\}$ CoT samples.  The shaped points are the other strategies from Section \ref{results:gsm8k} that use CoT and Subq (QG).%
    }\vspace{-12mm}
    \label{fig-sampcost}
\end{wrapfigure}

\our\ supports many methods with different cost/accuracy tradeoffs.  
Figure \ref{fig-sampcost} displays the strategies that use CoT and Subq (QG) on GSM8K. The cost is represented as the total count of input tokens (prompt + query) and output tokens for all LLM calls needed by that strategy, averaged over test examples. Generally, Subq (QG) is expensive as it is costly to call $\psi_{\text{question}}$.
However, it is affordable to use it in \textit{Conditional Resampling} only ($\textcolor{OliveGreen}{\blacksquare}$), since resampling only occurs 10--15\% of the time.  This method is both cheaper and more accurate than \textit{Sampling} either with Subq (QG) (\textcolor{DarkRed}{\bf +}) or 3 times with CoT ($\textcolor{Blue}{\bullet}$).
Appendix \ref{sampling-cost} discusses a detailed breakdown of each module's input and output token costs. 

\vspace{1mm}
\subsection{More Revision Steps}

We saw in Section \ref{results:gsm8k} 
on GSM8K that \textit{Sampling} with Subq (Or) (78.62 accuracy) is improved slightly by \textit{Conditional Resampling} with CoT (78.99) and then \textit{Selection} (79.22).  Like \citet{self-refine}, we did not find much benefit from additional iterations of \textit{Conditional Resampling}+\textit{Selection}: a second iteration gives 79.45,
and a third gives 79.52.  These small improvements probably do not justify the added cost.

\subsection{Larger LLMs}

Replacing ChatGPT with GPT-4 greatly increased the \textit{Sampling} accuracy on GSM8K, to 91.45 for CoT and 90.80 for Subq (Or). 
Choosing between those two samples with GPT-4-based \textit{Selection} further increased the accuracy to 93.10, which falls between the accuracy of majority voting over $k=3$ and $k=4$ CoT samples from GPT-4 (92.94 and 93.93 respectively). Even using ChatGPT-based \textit{Selection}  achieved 92.58, which is still an improvement over CoT alone. %

\subsection{Selected Examples}
The top two examples of Figure \ref{fig:qualitative}, on the GSM8K dataset, demonstrate the effectiveness of the \textit{Selection} module. The first example shows how an error introduced by \emph{Conditional Resampling} can be reverted by \emph{Selection}. The second example shows how a correction found by \emph{Conditional Resampling} can be kept by \emph{Selection}. 

The last example in Figure \ref{fig:qualitative}, on the StrategyQA dataset, illustrates that ordinary \textit{Resampling} is unlikely to correct an incorrect fact generated by the LLM. However, providing the correct facts during \textit{Resampling} gives the model access to new information, leading to the correct answer.

\begin{figure}[h!]
\scriptsize
\centering
\begin{tabular}{l}
\toprule
\addlinespace
\parbox{0.95\textwidth}{\textbf{Question} Gunter is trying to count the jelly beans in a jar. He asks his friends how many they think are in the jar. One says 80. Another says 20 more than half the first one. A third says 25\% more than the first one. What is their average guess?} \\
\addlinespace
\parbox{0.95\textwidth}{\textbf{Sampling} (A) The second person guesses  $20 + 1/2*80 = \texttt{<<}20+1/2*80=60\texttt{>>}60$ jelly beans. The third person guesses $80*1.25 = \texttt{<<}80*1.25=100\texttt{>>}100$ jelly beans. The total number of jelly beans is $80 + 60 + 100 = \texttt{<<}80+60+100=240\texttt{>>}240$. The average guess is $240/3 = \texttt{<<}240/3=80\texttt{>>}80$. The answer is 80.\textcolor{green}{\ding{51}} } \\
\addlinespace
\parbox{0.95\textwidth}{\textbf{Conditional Resampling} (B) The average guess is $(80 + 60 + 1.25*80)/3 = \texttt{<<}(80+60+1.25*80)/3=85\texttt{>>}85$ jelly beans. The second person guesses $80/2 + 20 = \texttt{<<}80/2+20=60\texttt{>>}60$ jelly beans. The average guess is $(80 + 60 + 1.25*80)/3 = \texttt{<<}(80+60+1.25*80)/3=85\texttt{>>}85$ jelly beans. The answer is 85. \textcolor{red}{\ding{55}}}\\
\addlinespace
\parbox{0.95\textwidth}{\textbf{Selection} \hllime{(A)} \textcolor{green}{\ding{51}} } \\

\addlinespace
\midrule
\addlinespace
\parbox{0.95\textwidth}{\textbf{Question} Adam has \$100 and wants to spend it to open a rock stand. He can buy rocks for \$5 each and sell them for \$7 each. If he invests all his money in the rock stand but only sells 60\% of his inventory, how much money does he lose?}\\
\addlinespace
\parbox{0.95\textwidth}{\textbf{Sampling} (A) Adam can buy \$100/\$5 per rock = $\texttt{<<}100/5=20\texttt{>>}20$ rocks. Adam sells 60/100 x 20 rocks = $\texttt{<<}60/100*20=12\texttt{>>}12$ rocks. Adam makes \$7 x 12 rocks = $\$\texttt{<<}7*12=84\texttt{>>}84$. Adam loses \$100 - \$5 x 20 rocks = $\$\texttt{<<}100-5*20=0\texttt{>>}0$. He invested all his money and sold all the rocks he bought, so he didn't lose any money. The answer is 0. \textcolor{red}{\ding{55}}}\\
\addlinespace
\parbox{0.95\textwidth}{\textbf{Conditional Resampling} (B) Adam can buy \$100/\$5 per rock = $\texttt{<<}100/5=20\texttt{>>}20$ rocks. Adam sells 60/100 x 20 rocks = $\texttt{<<}60/100*20=12\texttt{>>}12$ rocks. Adam makes \$7 x 12 rocks = $\$\texttt{<<}7*12=84\texttt{>>}84$. However, he spent \$5 x 20 rocks = $\$\texttt{<<}5*20=100\texttt{>>}100$, so he loses \$100 - \$84 = $\$\texttt{<<}100-84=16\texttt{>>}16$. The answer is 16. \textcolor{green}{\ding{51}} }\\
\addlinespace
\parbox{0.95\textwidth}{\textbf{Selection} \hllime{(B)} \textcolor{green}{\ding{51}} }\\
\addlinespace
\midrule \midrule
\addlinespace
\parbox{0.95\textwidth}{\textbf{Question}: Would Alexander Hamilton have known about koalas? }\\
\addlinespace
\parbox{0.95\textwidth}{\textbf{Sample}: Alexander Hamilton died in 1804. The first koala specimen was collected in 1792. It is possible that Hamilton knew about koalas, but it is unlikely. The answer is Unknown \textcolor{red}{\ding{55}}}\\
\addlinespace
\parbox{0.95\textwidth}{\textbf{Conditional Resampling}: Alexander Hamilton died in 1804. The first koala specimen was collected in 1792. It is possible that Hamilton knew about koalas. The answer is True \textcolor{red}{\ding{55}}}\\
\addlinespace
\parbox{0.95\textwidth}{\textbf{Facts}: \textit{Alexander Hamilton died in 1804. The first published depiction of a koala was in 1810}}\\
\addlinespace
\parbox{0.95\textwidth}{\textbf{Tool-based Resample}: \hllime{No, Alexander Hamilton would not have known about koalas. The answer is False}  \textcolor{green}{\ding{51}}}\\
\addlinespace
\bottomrule
\end{tabular}
\caption{The top two examples demonstrate the importance of the \emph{Selection} module for the GSM8K dataset. The last example shows how tool use (``Facts'') can be helpful for the StrategyQA dataset.}
\label{fig:qualitative}
\end{figure}

\section{Discussion}
\subsection{Key Findings}
Based on our experiments with three reasoning datasets using our framework, we conclude the following:
\begin{itemize}
    \item \textbf{\textit{Selection} plays an important role}: Although \textit{Conditional Resampling} often improves the result of \textit{Sampling},  \textit{Selection} can help avoid errors from the case where it does not.  It was beneficial on all three datasets.
    \item \textbf{Heterogeneous vs.\@ homogeneous resampling}: Using different reasoning methods for \textit{Sampling} and \textit{Conditional Resampling} can lead to higher accuracy, with or without \textit{Selection}. 
    \item \textbf{Missing external knowledge hurts \textit{Conditional Resampling}}: Resampling cannot fix incorrect facts generated by the model. Tool-based resampling can therefore get better results (as simulated using StrategyQA).
    \item \textbf{No uniformly best strategy}:  There was no clear winning method for each of the modules.  Simple baseline methods sometimes beat more complex ones: CoT uses only one call to $\psi$ and beats Subq (QG) in GSM8K, always selecting $\texttt{y}_\text{next}$ beats self-select for StrategyQA with ``Facts,'' and Answer Only works surprisingly well for Code Debugging. 
\end{itemize}

\subsection{Future Work}

\our~combines the three important modules \textit{Sampling}, \textit{Conditional Resampling} and \textit{Selection} in a modular framework. 
The best configuration of modules will vary by task and could be identified through a method such as exhaustive search, Monte Carlo Tree Search, or reinforcement learning.  The modules themselves could be fine-tuned to improve end-to-end performance.

If we want to optimize cost along with accuracy, \citep{chen2023frugalgpt} proposed several methods for speeding up the stochastic functions $\psi$.  Their ``LLM Cascade'' strategy in particular is a heterogeneous (but unconditional) resampling method that starts with smaller, cheaper models.

It is possible that for some reasoning tasks, additional modules could be useful.  For instance, \textit{Resampling} or \textit{Selection} might be preceded by \textit{Critiquing}, or \textit{Selection} might be generalized to \textit{Combination}.

\subsection{Conclusion}

We have proposed \our, a modular reasoning-with-revisions framework to answer reasoning questions with LLMs. We demonstrated the usefulness of the three main components of the framework---\textit{Sampling}, \textit{Conditional Resampling}, and \textit{Selection}---on three reasoning datasets. The flexible nature of our framework allows it to be configured for each task and extended to other tasks in the future. 

\newpage
\bibliography{iclr2024_conference}
\bibliographystyle{iclr2024_conference_url}

\newpage
\appendix

\section{Token Cost}
\label{sampling-cost}

Table \ref{sampling-cost} shows the token cost of input and output for each module in \our. Due to its iterative nature, subquestion decomposition requires on average four times more input tokens than the other modules. 
For \emph{Conditional Resampling}, the model first predicts whether it wants to modify its output or not, using one token (``Yes" or ``No") for each sample and then only for the answers starting with ``No", it resamples.  For the \emph{Selection} module, the model chooses one of the two samples presented to it, using one token (\texttt{A} or \texttt{B}) for the output.

\begin{table*}[h]
\centering
\begin{tabular}{l  c   c  c }
 \toprule
Method & Input Tokens & Output Tokens & Total Tokens \\
\midrule 
\addlinespace
\multicolumn{4}{c}{\textbf{Subquestion generation step} $\psi\textsubscript{question}$} \\
\midrule
Subq (QG) & 360 & 180 & 540   \\
\midrule
\multicolumn{4}{c}{\textbf{Sampling step} $\psi$} \\
\midrule 
CoT & 774 & 307 & 1081   \\  
CoT ($k=5$) & 3870 & 1530 & 5400 \\
Subq (Or) & 3187 & 413 & 3600 \\
Subq (QG) & 3121 & 434 & 3555   \\
\midrule 
\addlinespace
\multicolumn{4}{c}{\textbf{Conditional Resampling step} $\psi\textsubscript{ask}$ }\\
\midrule 
CoT & 869 & 105 & 1184   \\  
Subq (Or) & 3525 & 131 & 3656 \\
Subq (QG) & 3780 & 136 & 3916   \\
\midrule 
\multicolumn{4}{c}{\textbf{Selection step}  $\psi\textsubscript{select}$} \\
\midrule 
Selection & 1296 & 1 & 1297 \\
\bottomrule 
\end{tabular}
\caption{Average counts of input and output tokens for each choice of each module (step) in \our.   Many of the methods in Table~\ref{ResultsGSM} need to call multiple modules.
We remark that the input tokens at each step include output tokens from previous steps.  The counts shown for later steps average not only over examples, but also over choices of method for the previous steps.}
\label{cost}
\end{table*}

\section{Prompts}\label{app:prompts}

Below are abbreviated versions of the prompts used in the experiments, including instructions and demonstrations. For readability, we show only 1--2 demonstrations in each prompt. In each demonstration, the demonstrated result string is \hllime{highlighted} for the reader's convenience, but this highlighting is not included in the prompt.  Each prompt shown would be followed by the test question and then the cue (e.g., ``Answer:'') that indicates that a result string should follow.

\subsection{Sampling}
\label{sampling_prompts}
For Chain of Thought (CoT) and Subquestion Decomposition for GSM8K and StrategyQA, 5-shot prompts were used.  For Auto Debugging, a 1-shot prompt was used. 

\subsubsection{Chain of Thought}

\textbf{GSM8K} \\
\small
I am a highly intelligent question answering bot.  I will answer the last question `Question' providing equation in \texttt{<<} \texttt{>>} format in step by step manner.

Question: James writes a 3-page letter to 2 different friends twice a week.  How many pages does he write a year? \\
Answer: \hllime{He writes each friend $3*2=\texttt{<<}3*2=6\texttt{>>}6$ pages a week. So he writes $6*2=\texttt{<<}6*2=12\texttt{>>}12$ pages every week. That means he writes $12*52=\texttt{<<}12*52=624\texttt{>>}624$ pages a year. The answer is 624} \\
\normalsize

\hrule
\textbf{StrategyQA} \\
\small
You are a highly intelligent question answering bot. You will answer the question `Question' in as details as possible. \\
Question: Is coal needed to practice parachuting? \\
Answer: \hllime{Parachuting requires a parachute. Parachutes are made from nylon. Nylon is made from coal. The answer is True} \\
\normalsize
\hrule
\textbf{Auto Debugging} \\
\small
Answer the 'Question' based on the provided code and provide explanation.

Question:
\begin{minted}{python}
    def f1():
        return str(x) + 'hello'  
    def f2():
        return f1(2*x)
    x = f2(524)
\end{minted}   
What is the value of x at the end of this program? \\
Output: \hllime{First, x = 2 * 524 = 1048 and then `hello' is appended to it. So x becomes \texttt{1048hello} }\\
\normalsize
\subsubsection{Subquestion Decomposition}
While subquestion decomposition uses a single prompt, each example requires multiple API calls because the next subquestion needs to be appended to the prompt.

\textbf{GSM8K} \\
\small
I am a highly intelligent question answering bot.  I will answer the last question `Q' providing equation in \texttt{<<} \texttt{>>} format keeping the Problem and previous Q and A into account.

Problem: There are 5 houses on a street, and each of the first four houses has 3 gnomes in the garden. If there are a total of 20 gnomes on the street, how many gnomes does the fifth house have? \\
Q: How many gnomes are in the first four houses? \\
A: \hllime{In the first four houses, there are a total of 4 houses * 3 gnomes = $\texttt{<<}4*3=12\texttt{>>}12$ gnomes. The answer is 12} \\
Q: How many gnomes does the fifth house have? \\
A: \hllime{Therefore, the fifth house had 20 total gnomes - 12 gnomes = $\texttt{<<}20-12=8\texttt{>>}8$ gnomes. The answer is 8} \\
\normalsize
\hrule
\textbf{StrategyQA}\\
\small
You are a highly intelligent question answering bot. You will answer the last question `Q' keeping the Problem and previous Q and A into account and then answer the Final Question based on all the previous answer `A'.

Problem: Is coal needed to practice parachuting? \\
Q: What is one of the most important item that you need to go parachuting? \\
A: \hllime{Parachuting requires a parachute.} \\
Q: What is \#1 made out of? \\
A: \hllime{Parachutes are made from nylon.} \\
Q: Is \#2 originally made from coal? \\
A: \hllime{Nylon is made from coal.} \\
Final Question: Is coal needed to practice parachuting? \\
Final Answer: \hllime{True} \\
\normalsize
\hrule

\subsubsection{Answer Only}
Answer Only was only used for Auto Debugging in a 1-shot manner and that one example is provided below:

\small
Input:
\begin{minted}{python}
    def f1():
        return str(x) + 'hello'  
    def f2():
        return f1(2*x)
    x = f2(524)
\end{minted}   
What is the value of x at the end of this program? \\
Output: \hllime{\texttt{1048hello} }\\
\normalsize
\hrule

\subsection{Conditional Resampling}
\label{resampling_prompt}
We used 4-shot prompting for \emph{conditional resampling} (two positive and two negative) for GSM8K and StrategyQA and 2-shot (one positive and one negative) for Auto Debugging. One pair of each is presented below:

\subsubsection{Chain of Thought}

\textbf{GSM8K}\\
\small
Question: Natalia sold clips to 48 of her friends in April, and then she sold half as many clips in May. How many clips did Natalia sell altogether in April and May? \\
Answer: In April, Natalia sold 48 clips. In May, Natalia sold half of the toal clips meaning $72/2\texttt{<<}72/2=36\texttt{>>}36$ clips. So, Natalia sold a total of $48+36\texttt{<<}48+36=84\texttt{>>}84$ clips in April and May. The answer is 84 \\
You are a math teacher. Do you think the reasoning process for the given problem is correct? Let’s check the `Answer' in details, and then decide `Yes' or `No' and then write the correct `Final Answer'. \\
Answer: \hllime{No, the reasoning is incorrect. Final Answer: Natalia sold $48/2=\texttt{<<}48/2=24\texttt{>>}24$ clips in May. So she sold a total of $48+24\texttt{<<}48+24=72\texttt{>>}72$ clips. The answer is 72}

Question: There are 5 houses on a street, and each of the first four houses has 3 gnomes in the garden. If there are a total of 20 gnomes on the street, how many gnomes does the fifth house have? \\
Answer:In the first four houses, there are a total of 4 houses * 3 gnomes $= \texttt{<<}4*3=12\texttt{>>}12$ gnomes. Therefore, the fifth house had 20 total gnomes - 12 gnomes $= \texttt{<<}20-12=8\texttt{>>}8$ gnomes. The answer is 8 \\
You are a math teacher. Do you think the reasoning process for the given problem is correct? Let’s check the `Answer' in details, and then decide `Yes' or `No' and then write the correct `Final Answer'. \\
Answer: \hllime{Yes, the reasoning is correct. Final Answer: In the first four houses, there are a total of 4 houses * 3 gnomes $= \texttt{<<}4*3=12\texttt{>>}12$ gnomes. Therefore, the fifth house had 20 total gnomes - 12 gnomes  $ =\texttt{<<}20-12=8\texttt{>>}8$ gnomes. The answer is 8} \\
\normalsize
\hrule
\textbf{StrategyQA} \\
\small
Question: Could Durian cause someone's stomach to feel unwell? \\
Answer: Durian has a pungent odor that many people describe as being similar to feet and onions. Unpleasant smells can make people feel excited and they like it. The answer is False \\
You are an expert teacher. Do you think the reasoning process for the given problem is correct? Let’s check the `Answer' in details, and then decide `Yes' or `No' and then write the correct `Final Answer'. \\ 
Answer: \hllime{No, the reasoning is incorrect. Final Answer: Durian has a pungent odor that many people describe as being similar to feet and onions. Unpleasant smells can make people feel nauseous. The answer is True} \\

Question:  Was Daniel thrown into the lion's den in the New Testament? \\ 
Answer:The Book of Daniel is a book in the Old Testament of the Bible. The Bible is divided into the Old Testament and the New Testament. The New Testament focuses on four Gospels regarding the life of Jesus. The answer is False \\
You are an expert teacher. Do you think the reasoning process for the given problem is correct? Let’s check the `Answer' in details, and then decide `Yes' or `No' and then write the correct `Final Answer'. \\
Answer: \hllime{Yes, the reasoning is correct. Final Answer: The Book of Daniel is a book in the Old Testament of the Bible. The Bible is divided into the Old Testament and the New Testament. The New Testament focuses on four Gospels regarding the life of Jesus. The answer is False} \\
\normalsize
\hrule
\label{prompt:sqa_with_facts}
\textbf{StrategyQA (Resampling with facts)} \\
\small
You are a highly intelligent question answering bot. You will answer the question 'Question' in as details as possible. 'Facts' are provided to assist you in answering the questions. \\
Question: Are vinegar pickled cucumbers rich in lactobacillus? \\
Facts: Pickles made with vinegar are not probiotic and are simply preserved. Pickles made through a soak in a salt brine solution begin to ferment because of lactobacillus. \\
Answer: \hllime{No, vinegar does not contain lactobacillus. The answer is False} \\

Question: Does Masaharu Morimoto rely on glutamic acid?\\
Facts: Masaharu Morimoto is a Japanese chef. Japanese cuisine relies on several forms of seaweed as ingredients and flavorings for broth like kombu dashi. Glutamic acid has been identified as the flavoring component in kombu seaweed. \\
Answer: \hllime{Yes, Japanese chef uses a lot of glutamic acid. The answer is True}\\
\normalsize
\hrule
\textbf{Auto Debugging} \\
\small
Input:  \begin{minted}{python}
    def f1():
        return str(x) + 'hello'
    def f2():
        return f1(2*x)
    x = f2(524)
\end{minted}
What is the value of x at the end of this program? \\
Output: \hllime{\texttt{1048hello} \\
Verdict: Yes, the answer is correct. \\
Final Answer: \texttt{1048hello}} \\

Input:  \begin{minted}{python}
    def f1():
        return str(x) + 'hello'
    def f2():
        return f1(2*x)
    x = f2(524)
\end{minted}
What is the value of x at the end of this program? \\
Output: \hllime{\texttt{524} \\
Verdict: No, the answer is incorrect. \\
Final Answer: \texttt{1048hello}} \\
\normalsize
\hrule

\subsubsection{Subquestion Decomposition}

\textbf{GSM8K}\\
\small
For each subquestion, the main problem and all previous subquestions along with the model-generated solutions are provided in order to solve the current subquestion.\\

Here is a math question and its solution.\\
Problem: Noah is a painter. He paints pictures and sells them at the park. He charges \$60 for a large painting and \$30 for a small painting. Last month he sold eight large paintings and four small paintings. If he sold twice as much this month, how much is his sales for this month?\\
How much did Noah earn from the large paintings? Noah earned \$60/large painting x 8 large paintings = $\$\texttt{<<}60*8=480\texttt{>>}480$ for the large paintings. The answer is 480 \\
Question: How much did Noah earn from the small paintings? \\
Answer: He also earned \$60/small painting x 4 small paintings = $\$\texttt{<<}60*4=240\texttt{>>}240$ for the small paintings. The answer is 240 \\
You are a math teacher. Do you think the reasoning process for the given problem is correct? Let’s check the `Answer' in details, and then decide `Yes' or `No' and then write the correct `Final Answer'. \\
Answer: \hllime{No, the reasoning is incorrect. Final Answer: He also earned \$30/small painting x 4 small paintings = $\$\texttt{<<}30*4=120\texttt{>>}120$ for the small paintings. The answer is 120 }

Here is a math question and its solution. \\
Problem: To make pizza, together with other ingredients, Kimber needs 10 cups of water, 16 cups of flour, and 1/2 times as many teaspoons of salt as the number of cups of flour. Calculate the combined total number of cups of water, flour, and teaspoons of salt that she needs to make the pizza. \\
How many teaspoons of salt does Kimber need? To make the pizza, Kimber half as many teaspoons of salt as the number of cups of flour, meaning she needs 1/2*16 = $\texttt{<<}16*1/2=8\texttt{>>}8$ teaspoons of salt. The answer is 8 \\
How many cups of flour and teaspoons of salt does Kimber need? The total number of cups of flour and teaspoons of salt she needs is 8+16 = $\texttt{<<}8+16=24\texttt{>>}24$. The answer is 24 \\
Question: How many cups of water, flour, and salt does Kimber need? \\
Answer: She also needs 10 cups of water, which means the total number of cups of water and flour and teaspoons of salt she needs is $24+10=\texttt{<<}24+10=34\texttt{>>}34$. The answer is 34 \\
You are a math teacher. Do you think the reasoning process for the given problem is correct? Let’s check the `Answer' in details, and then decide `Yes' or `No' and then write the correct `Final Answer'. \\
Answer: \hllime{Yes, the reasoning is correct. Final Answer: She also needs 10 cups of water, which means the total number of cups of water and flour and teaspoons of salt she needs is $24+10=\texttt{<<}24+10=34>\texttt{>>}34$. The answer is 34} \\
\normalsize
\hrule
\textbf{StrategyQA} \\ 
\small
Here is a question and its answer. \\
Context: Would a diet of ice eventually kill a person? \\
Ice is the solid state of what? Ice can be melted into water, which consists of hydrogen and oxygen. \\
What nutrients are needed to sustain human life? Humans need carbohydrates, proteins, and fats that are contained in foods. \\
Question: Are most of \#2 absent from \#1? \\
Answer: Water does not contain fat, carbohydrates or protein. \\
You are an expert teacher. Based on the provided context, do you think the reasoning process for the given problem is correct? Let’s check the `Answer' in details, and then decide `Yes' or `No' and then write the correct `Final Answer'. \\
Answer: \hllime{Yes, the reasoning is correct. Final Answer: Water does not contain fat, carbohydrates or protein.} \\

Here is a question and its answer. \\
Context: Can binary numbers and standard alphabet satisfy criteria for a strong password? \\
Which characters make up binary numbers? Binary numbers only contain 0 and 1. \\
Which characters make up the standard English alphabet? The standard alphabet contains twenty six letters but no special characters. \\
Question: Does \#1 or \#2 include special characters or symbols? \\
Answer: Yes, it contains all the special characters. \\
You are an expert teacher. Based on the provided context, do you think the reasoning process for the given problem is correct? Let’s check the `Answer' in details, and then decide `Yes' or `No' and then write the correct `Final Answer'. \\
Answer: \hllime{No, the reasoning is incorrect. Final Answer: Neither binary digits nor English alphabets consists of any special characters which is needed for a strong password.} \\
\normalsize
\hrule

\subsection{Selection}
\label{selection_prompt}
The LLM-based selection module $\psi\textsubscript{select}$ uses a 2-shot prompt.  The 2 demonstrations in the prompt are shown below, for each dataset.

\textbf{GSM8K} \\
\small
You are an expert math teacher. You are provided with a question and two answers. Lets check the `Answer choices' step by step, and then decide which answer is correct `(A)' or `(B)' \\
Question: Natalia sold clips to 48 of her friends in April, and then she sold half as many clips in May. How many clips did Natalia sell altogether in April and May? \\
Answer choices: \\
(A) In April, Natalia sold 48 clips. In May, Natalia sold 24 clips. So, Natalia sold a total of 72 clips in April and May. The answer is 72. So in May she sold 48 clips. Total clips sold in April and May = $72 + 48 = \texttt{<<}72+48=120\texttt{>>}120$. The answer is 120 \\
(B) Natalia sold $48/2=\texttt{<<}48/2=24\texttt{>>}24$ clips in May. The answer is 24. Natalia sold $48+24=\texttt{<<}48+24=72\texttt{>>}$ clips altogether. The answer is 72 \\
Answer: (\hllime{B)} \\

You are an expert math teacher. You are provided with a question and two answers. Lets check the `Answer choices' step by step, and then decide which answer is correct `(A)' or `(B)' \\
Question: Dolly has two books. Pandora has one. If both Dolly and Pandora read each others' books as well as their own, how many books will they collectively read by the end? \\
Answer choices: \\
(A) There are a total of $2+1=\texttt{<<}2+1=3\texttt{>>}3$ books. The answer is 3. Dolly and Pandora both read all 3 books, so 3 books/person x 2 people = $\texttt{<<}3*2=6\texttt{>>}6$ books total. The answer is 6 \\
(B) The total number of books are $2 * 1 =\texttt{<<}2*1=2\texttt{>>}2$ books. The answer is 2. Dolly and Pandora read each other's books as well as their own, so the total number of books they read is 3 books. The answer is 3 \\
Answer: (\hllime{A)} \\
\normalsize
\hrule

\textbf{StrategyQA} \\
\small
You are the expert in the field. You are provided with a question and two answers. Lets check the reasoning process of each of the answer step by step, and then decide which answer is correct `(A)' or `(B)' \\
Question:  Could Durian cause someone's stomach to feel unwell? \\
Answer choices: \\
(A) Durian has a pungent odor that many people describe as being similar to feet and onions. Unpleasant smells can make people feel nauseous. The answer is True \\
(B) Durian has a pungent odor that many people describe as being similar to feet and onions. Unpleasant smells can make people feel excited and they like it. The answer is False \\
Answer: (\hllime{A)} \\

You are the expert in the field. You are provided with a question and two answers. Lets check the reasoning process of each of the answer step by step, and then decide which answer is correct `(A)' or `(B)' \\
Question: Was Daniel thrown into the lion's den in the New Testament? \\
Answer choices: \\
(A) The Book of Daniel is a book in the New Testament of the Bible. The Bible is divided into the Old Testament and the New Testament. The New Testament focuses on the life of Daniel. The answer is True \\
(B) The Book of Daniel is a book in the Old Testament of the Bible. The Bible is divided into the Old Testament and the New Testament. The New Testament focuses on four Gospels regarding the life of Jesus. The answer is False \\
Answer: (\hllime{B)} \\
\normalsize
\hrule

\textbf{Auto Debugging} \\
\small
You are an expert Python debugger. You are provided with a question and two answers. Your job is to decide which answer is correct `(A)' or `(B)' \\
Question: \begin{minted}{python}
def f1():
    return str(x) + 'hello'
def f2():
    return f1(2*x)
x = f2(524)    
\end{minted}
What is the value of x at the end of this program? \\
Answer choices: \\
(A) \texttt{524hello} \\
(B) \texttt{1048hello} \\
Answer: (\hllime{B)} \\
\normalsize
\hrule

\subsection{Question Generation}
\label{qg_prompt}
5-shot prompts were used for generating subquestions for GSM8K dataset. An example is provided below: \\

\textbf{GSM8K} \\
\small
I am a highly intelligent question generation bot.  I will take the given question `Q' and will decompose the main question into all `subquestions' required to solve the question step by step. \\

Q: James writes a 3-page letter to 2 different friends twice a week.  How many pages does he write a year? \\
Subquestions: \hllime{How many pages does he write each week? How many pages does he write every week? How many pages does he write a year?} \\
\normalsize
\hrule

\textbf{StrategyQA} \\
\small
I am a highly intelligent question generation bot.  I will take the given question `Q' and will decompose the main question into all `subquestions' required to solve the question step by step. \\

Q: Can you buy Casio products at Petco?\\
Subquestions: \hllime{What kind of products does Casio manufacture? What kind of products does Petco sell? Does \#1 overlap with \#2?}\\
\normalsize
\hrule

\end{document}

%% file: math_commands.tex
\usepackage{amsmath,amsfonts,bm}

\def\eqref#1{equation~\ref{#1}}

\def\1{\bm{1}}

\DeclareMathAlphabet{\mathsfit}{\encodingdefault}{\sfdefault}{m}{sl}
\SetMathAlphabet{\mathsfit}{bold}{\encodingdefault}{\sfdefault}{bx}{n}